\documentclass[runningheads]{llncs}
\usepackage[T1]{fontenc}
\usepackage{graphicx}
\usepackage{verbatim}

\usepackage[normalem]{ulem} 
\usepackage[table,xcdraw]{xcolor} 
\usepackage{multirow}
\usepackage{adjustbox}
\usepackage[table,xcdraw]{xcolor}
\usepackage{svg}
\usepackage{hyperref}

\usepackage{booktabs}

\newcommand{\textcode}[1]{\texttt{\small{#1}}}

\begin{document}
\title{Ontology Generation using\\ Large Language Models}
\def\thefootnote{*}\footnotetext{Equal contribution.}

\author{Anna Sofia Lippolis $^{1,2,*}$ 
\and
Mohammad Javad Saeedizade $^{3,*}$ 
\and \\
Robin Keskisärkkä \inst{3}
\and
Sara Zuppiroli\inst{2}
\and
Miguel Ceriani\inst{2}
\and
Aldo Gangemi \inst{1,2}
\and \\
Eva Blomqvist \inst{3}  
\and 
Andrea Giovanni Nuzzolese  \inst{2}  
}
\authorrunning{A. S. Lippolis et al.}

\institute{
  University of Bologna, Italy \\
  \email{annasofia.lippolis2@unibo.it, aldo.gangemi@unibo.it}
  \and
  ISTC-CNR, Italy \\
  \email{andrea.nuzzolese@cnr.it,sara.zuppiroli@istc.cnr.it,miguel.ceriani@cnr.it}
  \and
  Linköping University, Sweden \\
  \email{javad.saeedizade@liu.se, robin.keskisarkka@liu.se, eva.blomqvist@liu.se}
}

\maketitle
\begin{abstract}

The ontology engineering process is complex, time-consuming, and error-prone, even for experienced ontology engineers. In this work, we investigate the potential of Large Language Models (LLMs) to provide effective OWL ontology drafts directly from ontological requirements described using user stories and competency questions. Our main contribution is the presentation and evaluation of two new prompting techniques for automated ontology development: Memoryless CQbyCQ and Ontogenia. We also emphasize the importance of three structural criteria for ontology assessment, alongside expert qualitative evaluation, highlighting the need for a multi-dimensional evaluation in order to capture the quality and usability of the generated ontologies. Our experiments, conducted on a benchmark dataset of ten ontologies with 100 distinct CQs and 29 different user stories, compare the performance of three LLMs using the two prompting techniques. The results demonstrate improvements over the current state-of-the-art in LLM-supported ontology engineering. More specifically, the model \texttt{OpenAI o1-preview} with Ontogenia produces ontologies of sufficient quality to meet the requirements of ontology engineers, significantly outperforming novice ontology engineers in modelling ability. However, we still note some common mistakes and variability of result quality, which is important to take into account when using LLMs for ontology authoring support. We discuss these limitations and propose directions for future research.

\end{abstract}

\renewcommand*{\thefootnote}{\arabic{footnote}}
\setcounter{footnote}{0} 

\section{Introduction}
\label{sec:intro}
Ontologies play an important role in the success of Knowledge Graphs (KGs), as a crucial component in the recent advancements of explainable AI and neuro-symbolic integration. Today, ontologies are extensively used to facilitate semantic interoperability, e.g., describing datasets and standardised terminologies. However, ontology engineering (OE) is a complex task that requires skills in knowledge representation, logic, and computational linguistics. So far, many OE methodologies have emerged in the scientific literature to provide structured frameworks that assist ontology engineers in navigating the complexities of knowledge modeling. Examples are 
\textsc{Methontology}~\cite{Fernandez1997}, the NeOn methodology~\cite{Suárez-Figueroa2012}, eXtreme Design (XD)~\cite{10.1007/978-3-642-16438-5_9}, and more recently the Linked Open Terms \cite{POVEDAVILLALON2022104755}.
In any case, OE requires access to significant domain expertise combined with knowledge engineering and modelling skills, posing a significant barrier to entry for many professionals. 
Moreover, even when expertise is available, the creation, curation, and validation of ontological elements are complex tasks, which are cognitively costly and mostly manual. 
Instead, Large Language Models (LLMs) have proven to be able to assist humans in a variety of tasks, ranging from programming co-pilots to data cleaning and statistical analysis. These advancements highlight the growing need to benchmark common semantic web tasks from the perspective of LLMs, transforming this effort from an ideal into an essential requirement and emphasizing the critical importance of research in this area \cite{AlharbiBerardinisTamma2024,AllenGroth2024,RebboudLisenaTroncy2024,PluEscobarTrouillez2024,TsanevaHerwantoSabou2024}.
Furthermore, prompting techniques, defined as an approach to carefully formulating input prompts, can be leveraged to elicit the pre-trained knowledge of LLMs to perform specific tasks without additional training or fine-tuning. Advanced strategies such as decomposed prompting, where the task is split into several pieces, have been shown to enhance LLM performance across a wide range of tasks~\cite{khot2022decomposed}.
In this work, we assume that LLMs can also be beneficial for ontology design by reducing manual labour for experienced ontology engineers, as well as assisting novice ontology engineers.


Accordingly, the research questions driving this work are: (i) To what extent can LLMs be used to support the generation of ontologies that meet a predefined set of requirements? (ii) What evaluation criteria are suitable for evaluating LLM-generated ontologies? (iii) What are the strengths and weaknesses of ontologies generated using LLMs?
To answer these questions, we provide a framework to assess LLM-pipelines with prompting techniques focused on OE, as well as the LLM-generated minimal ontology modules themselves.
In particular, expanding previous research \cite{saeedizade2024navigating}, we leverage a dataset of ontology requirements and a set of reference minimal ontology modules to evaluate two prompting techniques for supporting OE: Memoryless CQbyCQ and Ontogenia\footnote{The supplementary materials are available at \url{https://github.com/dersuchendee/Onto-Generation}.}. We further evaluate the performance of these prompting techniques, using state-of-the-art LLMs, and propose a set of evaluation criteria for comparing and evaluating different pipeline setups. 


The paper is organized as follows: 
Section \ref{sec:related} reviews relevant literature; Section \ref{sec:preliminary} clarifies the terminology used in this paper; Section \ref{sec:methodology} details our research methodology; Section \ref{sec:eval} outlines the evaluation criteria used to assess our approach; Section \ref{sec:results} presents our findings with detailed analysis; Section \ref{sec:discussion} discusses the implications of the results. Finally, Section \ref{sec:conclusion} summarizes the main findings.


\section{Related Work}
\label{sec:related}
\label{sec:rel}
This section presents related work about ontology engineering methods, ontology generation methods, and prompting techniques for ontology generation.

\subsubsection{Ontology engineering methods.}
Several methodologies have been developed for ontology engineering, e.g., \textsc{Methontology}~\cite{Fernandez1997} and NeOn~\cite{NeOnBook}. Agile methodologies focusing on the re-use of Ontology Design Patterns~\cite{blomqvist2005patterns,gangemi2005ontology}, like those surveyed in \cite{blomqvist2016engineering,peroni2016simplified,shimizu_modular_2022}, have become increasingly popular, reflecting real-world needs by maximizing the cognitive soundness, logical correctness, and effectiveness of ontological artefacts. The LOT methodology~\cite{POVEDAVILLALON2022104755} was presented as a compilation of experiences from many projects, methodologies, and tools. While some methods have explicit tool support, e.g., NeOn~\cite{NeOnBook}, most are still entirely manual. Ontology engineers often must match the modelling problem to requirements, which is generally complex and time-consuming. 

\subsubsection{Ontology generation with LLMs}
With the advent of LLMs, much research has been focused on the possibility of generating ontologies, or their specific elements, from natural language. The increased interest in this area is exemplified by the announcement of several special tracks in the semantic web conferences dedicated to these issues\footnote{E.g. see ESWC \url{https://2024.eswc-conferences.org/}, EKAW \url{https://event.cwi.nl/ekaw2024/cfp.html} and ISWC \url{https://iswc2024.semanticweb.org/}}.  Despite LLMs' well-known drawbacks, such as hallucinations, studies show they have the potential to perform efficiently in numerous tasks, with only a handful of examples and a well-designed prompt \cite{ali2023performance,hanna2023comparative}. 
  In \cite{garijo2024llms} the authors survey the tasks currently addressed by approaches for LLM-supported ontology engineering, and while some approaches focusing on the actual tasks of generating the OWL files themselves have been proposed, this is still an area that needs much further investigation.
The LLMs4OL approach~\cite{giglou2023llms4ol} used LLMs to extract relations among ontology classes or instances, but only among entities, not addressing the complete generation of an ontology. 
Other preliminary work~\cite{mateiu2023ontology} has used fine-tuned GPT models to translate restricted natural language sentences into DL axioms. However, such specific sentences do not represent realistic ontology requirements or scenarios, as targeted in our work. 
Furthermore, tools are appearing to support the practical integration of OWL with LLMs, e.g. \cite{he24deeponto}, while not providing any specific guidance for the ontology generation task. One recent work similar to our proposed approach is the NeOn-GPT pipeline proposed by Fathallah et al. \cite{neongpt2024}. However, that work only evaluates one single complex prompting technique, and evaluates the pipeline only on one single ontology generation example. In LLMs4Life\cite{fathallah2024llms4life}, the NeOn-GPT approach is extended to the life sciences domain. However, implementation details remain unclear, especially for what concerns the evaluation criteria.  
In Lippolis et al. and Saeeizade and Blomqvist \cite{lippolisontogenia,saeedizade2024navigating}, different methods and ontology evaluation approaches for ontology generation from requirements have been tested, obtaining results which are at least comparable to the quality of human novice modellers: these two works are the starting point of our current investigation. 
Our multi-dimensional assessment implements the proposal of Rebboud et al., \cite{RebboudLisenaTroncy2024}, where ontology conceptualization is proposed to be evaluated according to ontology evaluation criteria such as accuracy, completeness, and conciseness of the generated ontology as well as logical consistency. Furthermore, the authors propose the only existing benchmark for this task. However, it comprises entire ontologies rather than minimal ontology modules, making it harder to evaluate the outputs LLMs at a wider granularity, and often lacks ontology stories.  

\subsubsection{Prompting techniques for ontology generation.}
Prompt engineering stands out for its simplicity in implementation and adaptability, in contrast to finetuning, offering an approach for enhancing knowledge engineering processes while avoiding the need for large labelled datasets and dedicated finetuned models. Requiring modest effort to implement, it provides substantial flexibility for updates and modifications. In Lippolis et al. and Saeeizade and Blomqvist \cite{lippolisontogenia,saeedizade2024navigating}, the authors have explored and evaluated various prompting techniques and identified several techniques that appear suitable for different aspects of ontology generation. 
Our research on the one hand explores subtask-decomposed prompting, i.e., in this case the promising CQbyCQ method from Saeedizade and Blomqvist\cite{saeedizade2024navigating}, as well another prompting technique based on Chain Of Thought (CoT) \cite{wei2022}. CoT is known for the specific ability to successfully improve the performance of different tasks and decreasing hallucinations~\cite{chu2023survey}. A particular case of CoT is Metacognitive Prompting (MP) \cite{lippolisontogenia}, inspired by human introspective processes. It encourages self-evaluation through the introduction of a series of steps, improving performance over CoT. Starting from the studies of Brown et al.,\cite{brown2020}, Wei et al.\cite{wei2022}, and Wang et al. \cite{wang2022,wang2023}, which demonstrated significant improvements in LLM response abilities when using CoT or, more effectively, MP, our study examines the performance of various prompting strategies in ontology generation, comparing CoT with subtask-decomposed prompting approaches.

\section{Preliminaries}
\label{sec:preliminary}
In this section, we clarify and define terminology that will be used throughout this paper. These definitions are essential for understanding the concepts discussed in subsequent sections and provide details on the specific ways in which we interpret and employ these notions within the context of our research. 

\noindent\textbf{Ontology.} In this work, ontology \( O \) is defined as a set of classes, object properties, data properties and axioms.

\noindent\textbf{Validation Competency Question.} As defined in Keet and Khan \cite{keet2024roles}, we can say a CQ is of the validation type if it ensures it adequately reflects the domain it represents by validating the ontology’s content while it aligns with its intended meaning and representation. For simplicity, we refer to this as CQ.

\label{pre:model}
\noindent\textbf{Modelled Competency Question.} Given an ontology \( O \) and a validation competency question \( CQ_i \), if \( O \) includes all necessary elements—such as classes or properties—needed to write a validation SPARQL query to verify \( CQ_i \) (CQ verification~\cite{blomqvist2012ontology}), we say that \( CQ_i \) is modelled in \( O \) regardless of the quality of the modelling, whether it follows good or bad modelling practices. 

\label{pre:super}
\noindent\textbf{Superfluous Element.} For an ontology \(O\), a set of competency questions and a set of validation SPARQL queries 
, 
if a named class, object property or data property is not used in any SPARQL queries, it is considered a superfluous element of \(O\). An example is in Appendix \ref{sec:Metrics}.

\noindent\textbf{Minimal Ontology Module.} Given an ontology \(O\) and a competency question \(CQ_i\), \(O_i\) is a minimal ontology module for \(CQ_i\) if we remove all superfluous elements of \(O\) with respect to \(CQ_i\).

\label{pre:minor}

\noindent\textbf{Minor Issue.} For an ontology \(O\) and a competency question \(CQ_i\), if \(O\) includes all necessary elements except for only one object property or only one data property, and adding this single element to O would make \(CQ_i\) modelled, this is considered a minor issue in modelling \(CQ_i\).

\section{Methodology}
\label{sec:methodology}

The initial phase of our work involved manually creating a benchmark dataset using available ontologies accompanied by their corresponding requirements for CQ verification~\cite{blomqvist2012ontology}. 
We introduced two methods for ontology generation, namely Independent Ontology Generation and Incremental Ontology Generation, and adapted existing prompting techniques in our experiments to guide LLMs in generating ontologies. In this study, we primarily employ \textit{GPT-4}, identified as the best-performing LLM in earlier comparative analyses~\cite{LLM4KGs2024,saeedizade2024navigating}. Additionally, we independently compare \textit{GPT-4} with \textit{OpenAI o1-preview} and \textit{Llama-3.1-405B-instruct-16b} using the dataset proposed in Saeedizade and Blomqvist~\cite{saeedizade2024navigating}.

\subsection{Dataset creation}
In order to evaluate the generated ontologies using CQ verification~\cite{blomqvist2012ontology}, a dataset of CQs and user stories was developed, along with their corresponding minimal ontology modules. The process involved selecting a set of ontologies for the study, extracting CQs and stories, and subsequently extracting modules from these chosen ontologies in order to have minimal units for easier evaluation and possible future extensions of the work.
The dataset combines two main types of sources: extracted datasets and manually created datasets. The extracted datasets consist of CQs and stories derived from existing resources, where superfluous elements were removed to identify minimal modules. In contrast, the manually created datasets were created as a part of a controlled module engineering process. In this case, we include the dataset present in \cite{saeedizade2024navigating} called ``SemanticWebCourse'', where master’s students with a computer science background but no prior ontology design experience were tasked with creating semantic web solutions. Each student group submitted an initial solution, revised it based on teacher feedback, and resubmitted a final version to pass the assignment, resulting in two distinct solutions per group. However, as previous work questioned the generalisability of the LLMs generating ontologies~\cite{saeedizade2024navigating}, in this work, we gave priority to different ontology sources, primarily real-world ones.

An ontology was included in the dataset according to the following criteria: The ontology includes (i) a set of competency questions and (ii) a set of corresponding user stories. These criteria were motivated by the eXtreme Design methodology~\cite{presutti2009extreme}, which takes into consideration CQs and user stories as fundamental building blocks of ontologies. As a result, we selected a total of ten ontologies, with 100 distinct CQs, and 29 different user stories from four real-life semantic web projects and three educational ones. More specifically, these ontologies have CQs assigned to specific minimal ontology modules with different user story distributions.  
\subsubsection{Minimal ontology module partitioning.}
The main goal of developing the dataset was to support the generation of ontologies using LLMs. The CQs and user stories provide inputs for prompting techniques in the ontology generation process. The minimal ontology modules contained in the datasets provide ontologists with a gold standard for the expected output of the LLMs and are used to assist the ontologists' assessment, proving useful to be used in future work to fine-tune LLMs. We followed a few steps to create minimal ontology modules for each CQ. First, we manually checked to see if there were duplicate CQs and removed them. Then, from the ontologies, we removed superfluous elements so each module contains the minimum necessary classes and properties to effectively model the CQ concerning the ontology story. Finally, these modules were cross-checked and evaluated independently by two of the authors of this study.

\subsubsection{Dataset composition.}

The dataset is composed of \textbf{simple} and \textbf{complex} CQs. These have been divided into four categories as defined in Saeedizade and Blomqvist \cite{saeedizade2024navigating}: Data Property Modelling (10 CQs) and Object Property Modelling (25 CQs) are Simple CQs, while Reification (62 CQs) and Restrictions (3 CQs) are Complex CQs. The CQs, stories and corresponding ontologies are taken from Polifonia \footnote{\url{https://polifonia-project.eu/}},
Onto-DESIDE \footnote{\url{https://ontodeside.eu}}, WHOW\footnote{\url{https://whowproject.eu/}}, IKS\footnote{\url{https://cordis.europa.eu/project/id/231527}} and the dataset used in Saeedizade and Blomqvist~\cite{saeedizade2024navigating} consisting of three ontology stories and 15 CQs each. As an additional baseline for the experiments in this paper we used the same set of student solutions of the ``SemanticWebCourse'' as in ~\cite{saeedizade2024navigating}. 
In Saeedizade and Blomqvist,~\cite{saeedizade2024navigating}, the baseline for evaluation was the first submission and the last submissions of student groups, and because it is the current state of the art in ontology generation, we use the same baseline in this paper for comparison. The intuition behind using this baseline is to compare the modelling abilities of LLMs to novice ontology engineers, with and without expert feedback\footnote{We did not compare the original Ontogenia paper as, given its shortcomings, the technique has been revisited for the decomposed prompting technique.}.

\subsection{Prompting Techniques}
In the study by Saeedizade and Blomqvist~\cite{saeedizade2024navigating}, the performance of CQbyCQ, a technique based on sub-task decomposition prompting~\cite{khot2022decomposed}, was found to be similar to that of students in ontology development. In this prompting technique, an LLM models only one CQ at a time, and the output is merged with the previously generated ontology. Due to its success, we incorporated two variations of this prompt to design our prompting techniques. Furthermore, in the study of Lippolis et al.\cite{lippolisontogenia}, Metacognitive Prompting has proven an effective technique, especially when Ontology Design Patterns were provided, to generate richer pattern-based ontology formalizations. The previous methodologies were therefore refined and tested on the benchmark dataset.

\paragraph{\textbf{Memoryless CQbyCQ.}}
Memoryless CQbyCQ processes one CQ at a time, utilizing the ontology story to guide the LLM with a prompt in generating an ontology model for each CQ. It then merges the outputs into a single ontology.
Memoryless CQbyCQ is a variation of CQbyCQ ~\cite{saeedizade2024navigating}  that does not provide the LLM with the current state of ontology development. Unlike the CQbyCQ method, the LLM does not have access to previously generated ontologies and other CQs while modelling a specific CQ, which reduces the input context size of the LLM by $\sim$60\%. In fact, in Saeedizade and Blomqvist~\cite{saeedizade2024navigating} it was shown that long context can result in distraction of the LLM. Hence, the intuition for removing the memory part is that slight overlaps between partial solutions will be easier to resolve and correct by even a novice ontology engineer, than completely irrelevant or inconsistent solutions, which could be the result of distracting the model. As a result, each CQ is independently modelled, and then all the resulting models, representing CQs, are merged at the end. This prompting technique guides an LLM, gives it an ontologist persona, and introduces Turtle syntax for defining classes, properties, reifications, and other ontology engineering features. It also includes a story section that outlines ontology requirements and a CQ, followed by common pitfalls in ontology development using LLMs, such as producing an empty output or engaging in conversation.

\paragraph{\textbf{Ontogenia technique.}}
Ontogenia was first
defined in \cite{lippolisontogenia} and has been further refined for this work.
Like the CQbyCQ method, this prompting technique guides an LLM by instructing it to be an ontology engineer, and defines basic guidelines for effective ontology formalization. The model processes one CQ at a time. When prompted to model a set of CQs, it models and merges the generated ontology at each step and provides it in the context. 
The inputs for the prompt are: a user story, which was not available in the previous work, a set of ODPs and possibly previous output. The main idea was to transpose the Metacognitive Prompting as described in five steps in Wang et al.~\cite{wang2023} with the XD methodology~\cite{10.1007/978-3-642-16438-5_9}, which required (i) the use of pre-selected CQs and user stories; (ii) selection, reuse, and integration of selected Ontology Design Patterns\footnote{\url{http://ontologydesignpatterns.org/}}; and (iii) iterative re-evaluation to verify coverage of initial requirements. In the initial stage, the LLM interprets and contextualizes requirements, identifying the context and breaking down the CQs into logical elements to support the systematic identification of classes and properties. The next phase involves reflecting on the CQs to extend the ontology by incorporating relevant rules and restrictions. Once the ontology is developed, the decision confirmation stage ensures the final output is validated with a clear explanation of the reasoning process. Finally, the LLM evaluates the ontology's reasoning, creating test cases with instances to validate the correctness of the generated ontology. The resulting procedure, mapped to the five MP steps is shown in the Github repository, along with more details about this technique.



\begin{figure}
    \centering
    \includegraphics[width=0.85\linewidth]{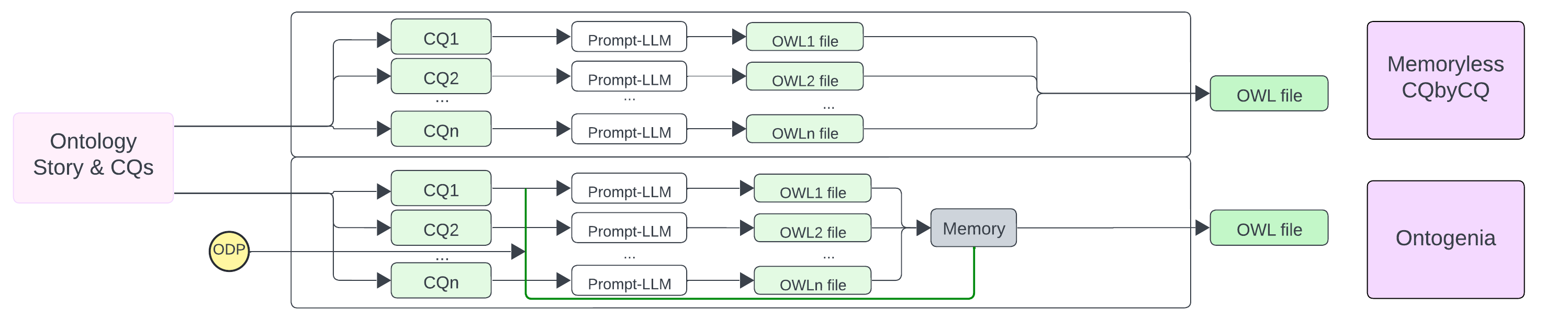}
    \caption{Illustration of Memoryless CQbyCQ (top part) and Ontogenia (bottom part).}
    \label{fig:prompt}
\end{figure}

\paragraph{\textbf{Similarities and differences between Memoryless CQbyCQ and Ontogenia.}}
Although the methods are similar in nature, and use the same testing setup
to ensure comparability, there are also some notable differences.
Ontogenia explicitly requests to provide labels, comments, inverse relationships and individuals. Furthermore, Ontogenia requires the injection of Ontology Design Patterns
from the ODP repository~\cite{blomqvist2005patterns} and makes use of the Metacognitive Prompting technique. Also, Ontogenia originally did not include scenarios in the prompt~\cite{lippolisontogenia}, unlike CQbyCQ~\cite{saeedizade2024navigating}. Unlike Ontogenia, Memoryless CQbyCQ includes common pitfalls to ensure they are avoided in the ontology output and attempts to reduce the context size. Figure~\ref{fig:prompt} shows an overview of these two prompting techniques.

\subsection{Ontology Generation Methods}

Here, we present the two ontology generation methods used in our experiments, as shown in Figure \ref{fig:expDiagram}: Independent and Incremental Ontology Generation.

\subsubsection{Independent Ontology Generation}
\label{singleCQ}
In this method, each CQ with its corresponding ontology story is fed into an LLM using a prompting technique to generate the corresponding ontology. 
This ensures that each CQ is treated as a standalone unit, allowing for a focused assessment of the ontology generation process on a per-question basis. By isolating each CQ, it becomes possible to analyze the performance of the prompting techniques without interference from interdependencies or complexities that may arise in multi-CQ contexts.



\subsubsection{Incremental Ontology Generation}
Here, similar to CQbyCQ ~\cite{saeedizade2024navigating}, all CQs of a story are fed to an LLM at once, and a single OWL ontology is expected as output.  Unlike the independent approach, where each CQ is modelled in isolation, this technique integrates solutions to multiple CQs to produce a cohesive minimal ontology module representing the complete story and its requirements. This method can be carried out either by merging the outputs for each CQ into a single ontology (Memoryless CQbyCQ) or by incorporating the output of each CQ incrementally into the prompt and joining it at each iteration (Ontogenia).





\section{Evaluation Setup}
\label{sec:eval}

\subsubsection{Experimental Setup.}
\begin{figure}
\label{fig:expDiagram}
    \centering
    \includegraphics[width=0.7\linewidth]{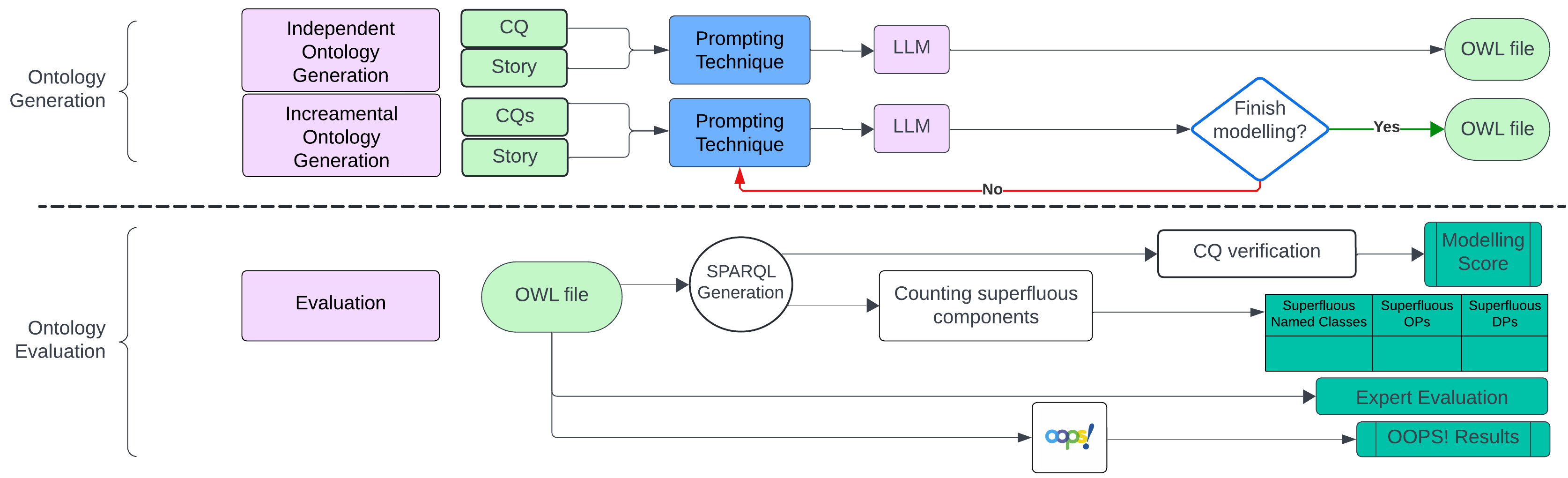}
\caption{Illustration of the two ontology generation settings (top) and the four evaluation steps for assessing the generated ontologies (bottom). The top setup generates an ontology concerning only one CQ, which is evaluated individually. The second setup generates an ontology covering multiple CQs associated with one story, which is then evaluated. At the bottom, the four ontology evaluation settings are shown: OOPS!, the proportion of modelled CQs, statistics of superfluous elements and expert evaluation.}
    
\end{figure}

The experiment consisted in running the Independent Ontology Generation method with \textit{GPT-4-1106} with both prompting techniques across the entire dataset. For the Incremental Ontology Generation method, we used both MemorylessCQbyCQ and Ontogenia with three different LLMs (\textit{GPT-4 1106}, \textit{o1-preview}, and \textit{Llama-3.1-405-Instruct-16b}) solely on the dataset introduced in Saeedizade and Blomqvist\cite{saeedizade2024navigating} of three semantic web course stories and 45 CQs.
As a baseline for comparison we used their results \cite{saeedizade2024navigating}, i.e. their best solutions produced using \textit{GPT-4} with CQbyCQ, and the recorded scores of the students' submissions. 
In order to provide a comprehensive evaluation, each of the solutions produced through these methods for the Independent and Incremental Ontology Generation experiments has been evaluated by the proportion of modelled CQs, with two ontology engineers cross-checking each other's judgments. In the event of any conflicting assessment, they engaged in discussions to resolve the disagreement. In the case of the Incremental Ontology Evaluation experiment, also standard ontology metrics through the OntOlogy Pitfall
Scanner (OOPS!) \cite{poveda2014oops}, as well as a structural analysis to evaluate the rate of superfluous elements and a qualitative expert evaluation, were carried out. Due to lack of resources and time, we were not able to perform this evaluation on the complete result set. For the LLMs, we used the default hyperparameters of \textit{o1-preview} and \textit{Llama-3.1-405-Instruct-16b}; for \textit{GPT-4}, temperature and penalty are set to zero as recommended in Saeedizade and Blomqvist~\cite{saeedizade2024navigating}. 
 Figure~\ref{fig:expDiagram} illustrates the three evaluation steps.


\subsubsection{Standard Ontology Metrics.}
The first step in evaluating the generated ontologies is applying standard evaluation metrics. We chose the OntOlogy Pitfall Scanner (OOPS!)
\cite{poveda2014oops}, due to its coverage of common modelling mistakes and best practices. In the results, we report only the number of critical issues related to each method, as they are the only ones crucial to correct, while other pitfalls are not certain to actually represent modelling flaws in the domain \cite{poveda2014oops}. 

\textbf{Proportion of Modelled CQs.}
In this step, for each CQ, we evaluate whether it is modelled in accordance with the definition provided in Section \ref{pre:model}. Additionally, we adopt a relaxed interpretation of this criterion by disregarding minor issues, as outlined in Section \ref{pre:minor}. Finally, the proportion of CQs deemed to be modelled, both with and without accounting for minor issues, is used to compute an overall score representing the proportion of the total list of CQs being sufficiently covered.

\textbf{Structural Analysis.}
\label{sec:structural}
We manually evaluated each OWL file using three novel criteria with respect to superfluous elements \ref{pre:super}, i.e., superfluous named classes, object properties, and data properties. We reported the rate of superfluous elements by dividing the number of superfluous elements by the total of that element and compared them to the rates in Saeedizade and Blomqvist~\cite{saeedizade2024navigating}.


\textbf{Expert Qualitative Analysis.}
The expert qualitative analysis was conducted by two experienced knowledge engineers, not involved in the practical ontology generation task, who carefully and independently evaluated the outputs generated by the best performing open- and closed-source LLMs, \textit{Llama-3.1-405-Instruct-16b}, and \textit{OpenAI o1-preview}. The experts were instructed to assess the LLM-generated ontology files as if they were students' course submissions and to provide feedback accordingly. Therefore, they focused on: (i) assessing the general quality of the output with dedicated comments on usability, completeness, and accuracy of the produced output, and mentioning any ontology errors found; and (ii) assessing the ontology with respect to the adequacy of the modelling solutions for each CQ\footnote{The ``not adequate'' assessment combines two judgments by the KE experts, i.e. a clear ``no'' where the CQ is not modelled, and a ``maybe'' category where the ontology simply does not allow for accurate assessment of the CQ, for instance, due to usability issues, naming etc.}, with a substantial agreement (Cohen's kappa: 0.61). 

\section{Results}
\label{sec:results}
In this section, we start by examining the results of the OOPS! resource on the incrementally generated ontologies. Then, we compare the two prompting techniques introduced in this paper by going in-depth into the modelling results with respect to the assessment of the proposed ontology generation methods.  
More specifically, we measure the proportion of modelled CQs for all experiments and analyze the structure of the generated ontologies concerning superfluous elements. For the Incremental Ontology Generation method, we focus on the ``SemanticWebCourse'' dataset since it contains 15 CQs associated with one story while others have $\sim$2 on average and choose the best performing open- and closed-source LLMs to compare across both previous work \cite{saeedizade2024navigating} and the results of Independent CQ Generation. In this way, we can test the LLMs' capability to generate an ontology incrementally.
Finally, we present the results of the assessment by the two KE experts who analyzed the outputs.

\paragraph{\textbf{Standard Ontology Metrics.}}

After running OOPS! on the ontologies generated by the Incremental method for the ``SemanticWebCourse'' stories (Hospital, Music, and Theatre), the resulting pitfalls are shown in Table~\ref{tab:my-tableoops1}. Overall, there are a few critical pitfalls for specific combinations of LLMs and prompting techniques. The most common flaw is having multiple domains or ranges, which is in line with our other results discussed further in the following sections (see Section \ref{sec:discussion}). In this case, the domain or range (or both) of a property (object or data property) is defined by more than one \textcode{rdfs:domain} or \textcode{rdfs:range} axiom. In OWL, multiple \textcode{rdfs:domain} or \textcode{rdfs:range} axioms are allowed, but are interpreted as a conjunction. Therefore, they are equivalent to the construct \textcode{owl:intersectionOf}, which could generate many unwanted inferences or even inconsistencies in the ontology. Some issues with inverse relations are also noted, as well as missing declarations of namespaces. Lastly, \textit{OpenAI o1-preview} produced the fewest critical issues when using Memoryless CQbyCQ, whereas \textit{Llama-3.1-405-instruct-16b} generated, with both methods, files with the most critical pitfalls.

\begin{table}[]
\caption{Pitfalls after OOPS! pitfall scanning for Incremental Ontology Generation. Results are merged by the three stories in the ``SemanticWebCourse''. Llama* refers to \textit{Llama-3.1-405-instruct-16b}. For the CQbyCQ method, only GPT-4 was used \cite{saeedizade2024navigating}.}
\centering
\label{tab:my-tableoops1}
\begin{adjustbox}{width=.78\textwidth}
\begin{tabular}{ll|c|ccc|ccc|}
\cline{3-9}
                          &                               & \multicolumn{1}{c|}{CQbyCQ}         & \multicolumn{3}{c|}{MemorylessCQbyCQ}   & \multicolumn{3}{c|}{Ontogenia}          \\ \cline{3-9} 
                          &                               & \multicolumn{1}{c|}{GPT-4}          & GPT-4 & Llama* & o1   & GPT-4& Llama* & o1   \\ \hline
\multicolumn{1}{|l|}{P05} & Wrong inverse relationships   & \textbf{0}                          & 1                        & 25     & \textbf{0} & 2                        & 7      & 5    \\ \hline
\multicolumn{1}{|l|}{P06} & Cycles in a class hierarchy   & \textbf{0}                          & \textbf{0}               & 2      & \textbf{0} & 5                        & \textbf{0} & 11   \\ \hline
\multicolumn{1}{|l|}{P19} & Multiple domains or ranges    & \textbf{0}                          & 23                       & 32     & 1    & 4                        & 15     & \textbf{0} \\ \hline
\multicolumn{1}{|l|}{P29} & Wrong transitive relationship & \textbf{0}                          & \textbf{0}               & \textbf{0} & 1    & \textbf{0}               & \textbf{0} & \textbf{0}    \\ \hline
\multicolumn{1}{|l|}{P37} & Ontology not available        & 3                                   & 2                        & \textbf{0} & \textbf{0} & \textbf{0}               & \textbf{0} & \textbf{0}    \\ \hline
\multicolumn{1}{|l|}{P39} & Ambiguous namespace           & 3                                   & 2                        & \textbf{0} & 1    & 1                        & \textbf{0} & \textbf{0}    \\ \hline
\end{tabular}

\end{adjustbox}
\end{table}

\subsubsection{Independent Ontology Generation Results.}
\label{sec:singleCQ}
The evaluation of the Memoryless CQbyCQ and Ontogenia techniques on \textit{GPT-4} shows that the proportion of correctly modelled CQs was 0.91 and 0.84, respectively (0.94 and 0.89 by ignoring minor issues), with significantly lower scores for complex CQs (0 and 0.66 respectively). These results show that by overlooking minor issues both prompting techniques are effective for modelling single CQs in isolation.

\subsubsection{Incremental Ontology Generation Results.}
\label{sec:multiCQ}
\begin{figure}[h]
\centering

\includegraphics[width=0.8\textwidth,height=0.25\textheight,keepaspectratio]{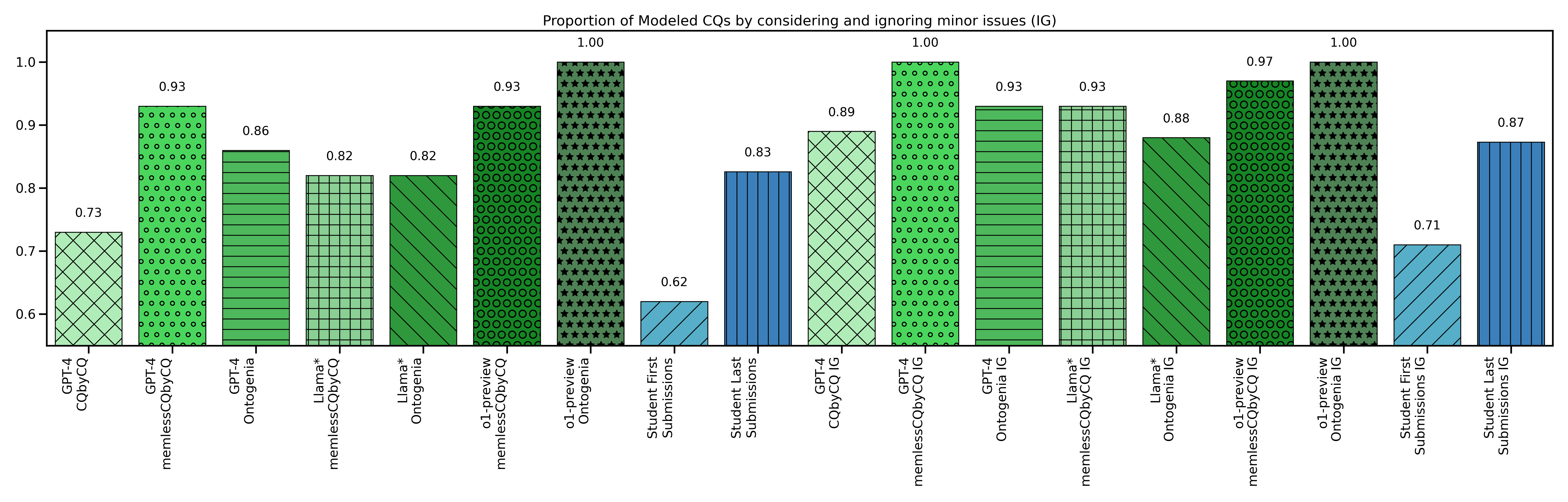
}
\caption{
Scores for ``SemanticWebCourse'' from the outputs for the different prompting techniques compared with students' submissions according to the proportion of the CQs that were accurately modelled. `IG' indicates results when minor issues are ignored. Llama$\star$ refers to \texttt{Llama-3.1-405B-instruct-bf16}.
}
\label{fig:image1}
\end{figure}

The results of the Incremental Ontology Generation are presented in Figure~\ref{fig:image1}.
Ontogenia with \textit{ OpenAI o1-preview} is the best-performing combination with respect to this evaluation criterion. Compared to the students' submissions for the ``SemanticWebCourse'', Ontogenia and Memoryless CQbyCQ with any LLM exceeded the proportion of CQs being modelled by students by a noticeable margin. Also, Memoryless CQbyCQ performed better than the CQbyCQ technique in Saeedizade and Blomqvist\cite{saeedizade2024navigating}, possibly suggesting that using partially generated ontology models as data ``in memory'' for LLMs, which includes previously generated responses to CQ modelling, actually may diminish their performance. This result highlights the importance of reducing the input context size for LLMs, which is consistent with the findings in Saeedizade and Blomqvist\cite{saeedizade2024navigating}. The proportion of correctly modelled CQs is analyzed across four categories: datatype property (DP), object property (OP), reification (Reif.), and restrictions (Rest.) for the Theatre, Music, and Hospital Story. We compared \textit{GPT-4}, \textit{Llama-3.1-405B-instruct-bf16} and \textit{OpenAI o1-preview}. The evaluation shows that Ontogenia with \textit{OpenAI o1-preview} achieves the highest number of correctly modelled CQs. Overall, both Ontogenia and Memoryless CQbyCQ surpass the students' last submissions and the state of the art in ontology generation \cite{saeedizade2024navigating}. Moreover, \textit{Llama-3.1} in both prompting techniques had shortcomings in modelling reification CQs.




\subsubsection{Structural Analysis.}
Table~\ref{tab:results2} presents our structural analysis regarding the superfluous elements. The results, compared to those in Section~\ref{sec:multiCQ}, show that the Memoryless CQbyCQ and Ontogenia, despite their performance measured in other criteria, yield a significant number of superfluous elements within the ontology. This underscores the importance of applying additional criteria for evaluating ontologies constructed using LLMs. CQbyCQ with \textit{GPT-4} produces the fewest superfluous elements compared to our newly introduced prompting techniques, while the Memoryless CQbyCQ approach yields comparable results. Conversely, \textit{Llama-3.1-405B-instruct-bf16} generates numerous superfluous elements across both techniques and all stories, with a rate close to 40\%, indicating that this model tends to produce many superfluous classes and properties.

\begin{table}[]
\small

\caption{Comparison of the three prompting techniques, using the Incremental Ontology Generation method, for generating superfluous elements across different ontology domains (Theatre, Music, and Hospital). Each cell represents the proportion of superfluous elements relative to the total number of elements of that type (\%). CQbyCQ was only tested with GPT-4 due to its low CQ-coverage. Rates under 15\% are bolded (best performance), and rates over 50\% are in red (worst performance).}

\label{tab:results2}
\centering
\begin{adjustbox}{width=0.88\textwidth}
\begin{tabular}{l|ccc|ccc|ccc}
\multicolumn{1}{c|}{\cellcolor[HTML]{FFFFFF}}                                      & \multicolumn{3}{c|}{Superflu. Classes}                                                        & \multicolumn{3}{c|}{Superflu. Obj. Properties}                                              & \multicolumn{3}{c}{Superflu. Data Properties}                                              \\ \cline{2-10} 
\multicolumn{1}{c|}{\multirow{-2}{*}{\cellcolor[HTML]{FFFFFF}Prompting  Technique}} & \multicolumn{1}{c|}{Theatre}    & \multicolumn{1}{c|}{Music} & Hospital & \multicolumn{1}{c|}{Theatre}    & \multicolumn{1}{c|}{Music} & Hospital & \multicolumn{1}{c|}{Theatre} & \multicolumn{1}{c|}{Music} & Hospital \\ \hline

CQbyCQ (GPT-4)  & \multicolumn{1}{c|}
{\textbf{0}}
& \multicolumn{1}{c|}
{{14.3}}
&\textbf{8.6}&

\multicolumn{1}{c|}
{\textbf{0}}
& \multicolumn{0}{c|}
{\textbf{4.2}}
&
\textbf{6.6}
& 

\multicolumn{1}{c|}
{\textbf{12.5}}
& \multicolumn{1}{c|}
{28.5}
&18
\\ \hline

Ontogenia (GPT-4)                     & \multicolumn{1}{c|}
{\textbf{0}}
& \multicolumn{1}{c|}
{19.2}
&      
\textbf{13.5}

& \multicolumn{1}{c|}
{\textbf{4}}
& \multicolumn{1}{c|}
{38.9}
&     
29.6

& \multicolumn{1}{c|}
{45.5}
& \multicolumn{1}{c|}
{\textcolor{red}{55.5}}
&         
22.2
\\ \hline

Ontogenia (Llama)  & 

\multicolumn{1}{c|}{47.1}
& \multicolumn{1}{c|}{32.1}
&37.5

& \multicolumn{1}{c|}
{40.6}
& \multicolumn{1}{c|}
{45.9}
& 
38.3

&\multicolumn{1}{c|}
{16.7}
& \multicolumn{1}{c|}
{\textbf{0}}
& \textcolor{red}{100}
\\ \hline

Ontogenia (o1)  & 

\multicolumn{1}{c|}{16.7}
& \multicolumn{1}{c|}{\textbf{12.5}}
&27.3

& \multicolumn{1}{c|}
{29.4}
& \multicolumn{1}{c|}
{55.2}
& 
37

&\multicolumn{1}{c|}
{28.6}
& \multicolumn{1}{c|}
{\textcolor{red}{66.7}}
& 45.5
\\ \hline

MemorylessCQbyCQ(GPT-4)  & 

\multicolumn{1}{c|}{25.7}
& \multicolumn{1}{c|}{31.2}
&38.9

& \multicolumn{1}{c|}
{17.4}
& \multicolumn{1}{c|}
{20.9}
& 
16.2

&\multicolumn{1}{c|}
{48}
& \multicolumn{1}{c|}
{40.9}
&   
46.1
\\ \hline

MemorylessCQbyCQ(Llama)  & 

\multicolumn{1}{c|}{45.6}
& \multicolumn{1}{c|}{42.3}
&23.9

& \multicolumn{1}{c|}
{20.5}
& \multicolumn{1}{c|}
{25.6}
& 
41.7

&\multicolumn{1}{c|}
{\textcolor{red}{60}}
& \multicolumn{1}{c|}
{\textcolor{red}{60.7}}
& \textbf{13.6}
\\ \hline

MemorylessCQbyCQ(o1)  & 

\multicolumn{1}{c|}{29.3}
& \multicolumn{1}{c|}{\textcolor{red}{50}}
&28.6

& \multicolumn{1}{c|}
{\textbf{14.8}}
& \multicolumn{1}{c|}
{\textbf{0}}
& 
\textbf{7.4}

&\multicolumn{1}{c|}
{\textbf{10}}
& \multicolumn{1}{c|}
{\textbf{0}}
& 28.6

\end{tabular}
\end{adjustbox}

\end{table}


\subsubsection{Expert Qualitative Analysis.}
After analyzing the results presented in Tables~\ref{tab:my-tableoops1} and \ref{tab:results2} and bar charts in Figure~\ref{fig:image1}, we conclude that \textit{OpenAI o1-preview}, especially with Ontogenia, seems to produce less errors overall than \textit{GPT-4}, and provides a better trade-off between different error types, notwithstanding all the generated files showed incorrect domains and ranges. Therefore, \textit{Llama-3.1-405B-instruct-bf16} and \textit{OpenAI o1-preview} are confirmed as the best open-source and closed-source LLMs, respectively, across both previous experiments \cite{saeedizade2024navigating} and the ones proposed in this work. The former's performance was not impressive: the ontologies generated by \textit{Llama-3.1-405B-instruct-bf16} show structural flaws, including inconsistent naming, redundant classes, and overlapping domains/ranges. The taxonomy shows circular references and poor property organization. Key issues include malformed cardinality restrictions, lack of comments/labels, and misaligned namespaces. Apart from superfluous elements, flat property hierarchies, and inconsistent axiomatisation, are visible. Table \ref{tab:qualanalysis} shows, for each selected story, the resulting percentage scores from the adequacy assessment of the qualitative evaluation. Different results are calculated as an average between the two. The full qualitative analysis can be seen on Github.

\begin{table}[ht]
\renewcommand{\arraystretch}{0.5} 
\setlength{\tabcolsep}{8pt} 
\centering
\small 
\caption{Adequate CQ modelling (\%) by Llama and o1 models for the selected stories.}
\label{tab:qualanalysis}
\begin{adjustbox}{width=.58\textwidth}

\begin{tabular}{l|l|c|c}

\toprule
\small

\textbf{Story} & \textbf{Model} & \textbf{Llama-3.1-405} & \textbf{o1-preview} \\ \midrule
\small

\multirow{2}{*}{\textbf{Music}} & MemorylessCQbyCQ       & 0.6   & 0.9   \\
                                & Ontogenia & 0.86  & 0.96  \\ \midrule
\multirow{2}{*}{\textbf{Theatre}} & MemorylessCQbyCQ        & 0.66  & 0.73  \\
                                  & Ontogenia & 0.63  & 1.0   \\ \midrule
\multirow{2}{*}{\textbf{Hospital}} & MemorylessCQbyCQ        & 0.63  & 0.73  \\
                                  & Ontogenia & 0.66  & 1.0   \\ 
\bottomrule
\end{tabular}
\end{adjustbox}

\end{table}

\section{Discussion}
\label{sec:discussion}
In this section, we present a discussion of the results.

\textbf{Overall results.}
According to the findings, and similarly to previous work~\cite{saeedizade2024navigating}, the use of an LLM yields promising results in supporting ontology engineering processes. In the Independent Ontology Generation, Memoryless CQbyCQ and Ontogenia perform well with Single Data Property CQs and Single Object Property CQs, but worse on generating Reifications and Restrictions. For the Incremental Ontology Generation method, Ontogenia and Memoryless CQbyCQ outperformed previous methods, including the students' solutions, with Ontogenia combined with \textit{OpenAI o1-preview} yielding the highest percentage of modelled CQs. However, the structural analysis reveals that with respect to previous work, the two prompting techniques generate more superfluous elements and yield critical pitfalls according to OOPS!. 
These structural issues are also noted by the experts in their evaluation, which found many flaws in the Llama-generated ontologies related to the usability and understandability of the solutions, while o1-generated ontologies are overall comparable to a student in the case of MemorylessCQbyCQ and even better in the case of Ontogenia. 

\textbf{Multidimensional evaluation.} Our multidimensional evaluation demonstrates that expert assessments align closely with both the standard and structural metrics used in this study. This suggests that experts, perhaps implicitly, rely on similar criteria when evaluating ontologies, resulting in evaluations that comprehensively address most aspects. At the same time, we still need qualitative evaluation for a holistic assessment of LLM-generated ontologies.

\textbf{Superfluous elements.}
An issue noted in LLM-generated ontologies involves the creation of superfluous properties with the same domain and range or multiple classes and properties that could be considered equal. For example, \texttt{employedSince} and \texttt{employmentStartDate} are generated for the same CQ, but only one is needed. This raises questions about superfluous elements: their number, why they are generated by the LLM, and their consequences. At the same time, having superfluous elements (in moderate numbers) may be considered less important than having an unmodeled CQs or more complex errors. Thus, our prompting techniques can be considered more complete and more usable than previous work  \cite{saeedizade2024navigating}, which was less accurate (Table \ref{tab:results2}), but more concise. In our envisioned setting of a future ontology engineering co-pilot, superfluous elements can be identified, e.g. through OOPS!, and manually removed or avoided with an even more elaborate prompt. As our approach aims to assist rather than replace ontology engineers, we believe this is acceptable.

\section{Conclusion}
\label{sec:conclusion}
In this paper, we introduced two novel and improved prompting techniques for ontology generation, Memoryless CQbyCQ and Ontogenia, assessing them through a multi-dimensional evaluation and with a new dataset with respect to previous studies. The results show the proposed prompting techniques proved promising to support the generation of ontologies that meet a predefined set of requirements, improving the proportion of modelled CQs, surpassing previous approaches and novice ontology modellers. However, challenges like multiple domains or ranges and other pitfalls were highlighted both by the OOPS! pitfall scanner and the expert evaluation. Both prompting techniques often generated superfluous elements, which (in low numbers) are not detrimental to ontology usability, compared to other major errors involving wrong axiomatisation and mistakes in the taxonomy. Returning to our research questions, regarding (i) to what extent LLMs can be used to support the generation of ontologies that meet a predefined set of requirements, we conclude that current commercial models, such as \textit{OpenAI o1}, can certainly perform on par with non-experts, at least when using a carefully selected prompt. Regarding (ii) what evaluation criteria are suitable for evaluating LLM-generated ontologies, we conclude that it is not sufficient to use one single criteria, nor simple automated evaluation metrics to detect errors and compare performance. Instead, rather elaborate manual methods are needed, e.g. to detect the amount of superfluous classes and properties. Finally, regarding (iii) what the strengths and weaknesses of ontologies generated using LLMs are, we believe that if cost is not an issue and commercial models can be used, then certainly novice ontology engineers could benefit from such draft ontologies generated by those LLMs. These drafts can increase modelling quality, and considerably reduce the effort and time to kick-start the modelling. However, certain weaknesses are observed across all the experiments, such as erroneous domain and range restrictions, erroneous inverse property axioms, and superfluous or overlapping classes and properties.


\begin{credits}
\subsubsection{\ackname}
This project has received funding from the European Union’s Horizon Europe research and innovation programme under grant agreements no. 101058682 (Onto-DESIDE) and 101070588 (HACID), and is supported by the strategic research area Security Link. The student solutions used in the research were collected as part of a master's course taught by Assoc. Prof. Blomqvist while employed at Jönköping University.
Additional financial support to this project was provided by NextGenerationEU under NRRP Grant agreement n. MUR IR0000008 - FOSSR (CUP B83C22003950001).
This work was also supported by the PhD scholarship ``Discovery, Formalisation and Re-use of Knowledge Patterns and Graphs for the Science of Science'', funded by CNR-ISTC through the WHOW project (EU CEF programme - grant agreement no. INEA/CEF/ICT/ A2019/2063229). Finally we thank OpenAI's Researcher Access Program Grant for the API credits.

\subsubsection{\discintname}
The authors have no competing interests to declare that are
relevant to the content of this article.
\end{credits}
%
%
%
\bibliographystyle{splncs04}
\bibliography{bibliography}

\begin{thebibliography}{10}
\providecommand{\url}[1]{\texttt{#1}}
\providecommand{\urlprefix}{URL }
\providecommand{\doi}[1]{https://doi.org/#1}

\bibitem{AlharbiBerardinisTamma2024}
Alharbi, R., de~Berardinis, J., Grasso, F., Payne, T., Tamma, V.: Characteristics and desiderata for competency question benchmarks. In: The Semantic Web – ISWC 2024: 23rd International Semantic Web Conference, Baltimore, MD, USA, November 11–15, 2024, Proceedings (2024)

\bibitem{ali2023performance}
Ali, R., Tang, O.Y., Connolly, I.D., Fridley, J.S., Shin, J.H., Zadnik~Sullivan, P.L., Cielo, D., Oyelese, A.A., Doberstein, C.E., Telfeian, A.E., Gokaslan, Z.L., Asaad, W.F.: Performance of chatgpt, gpt-4, and google bard on a neurosurgery oral boards preparation question bank. Neurosurgery  \textbf{93}(5),  1090--1098 (2023). \doi{10.1227/neu.0000000000002551}

\bibitem{AllenGroth2024}
Allen, B., Groth, P.: A benchmark for the detection of metalinguistic disagreements between llms and knowledge graphs. In: The Semantic Web – ISWC 2024: 23rd International Semantic Web Conference, Baltimore, MD, USA, November 11–15, 2024, Proceedings (2024)

\bibitem{giglou2023llms4ol}
Babaei~Giglou, H., D'Souza, J., Auer, S.: Llms4ol: Large language models for ontology learning. In: Payne, T.R., Presutti, V., Qi, G., Poveda-Villal{\'o}n, M., Stoilos, G., Hollink, L., Kaoudi, Z., Cheng, G., Li, J. (eds.) The Semantic Web -- ISWC 2023. pp. 408--427. Springer Nature Switzerland, Cham (2023)

\bibitem{balloccu2024leakcheatrepeatdata}
Balloccu, S., Schmidtov{\'a}, P., Lango, M., Dusek, O.: Leak, cheat, repeat: Data contamination and evaluation malpractices in closed-source {LLM}s. In: Graham, Y., Purver, M. (eds.) Proceedings of the 18th Conference of the European Chapter of the Association for Computational Linguistics (Volume 1: Long Papers). pp. 67--93. Association for Computational Linguistics, St. Julian{'}s, Malta (Mar 2024), \url{https://aclanthology.org/2024.eacl-long.5}

\bibitem{blomqvist2016engineering}
Blomqvist, E., Hammar, K., Presutti, V.: Engineering ontologies with patterns-the extreme design methodology. In: Ontology Engineering with Ontology Design Patterns. IOS Press (2016)

\bibitem{10.1007/978-3-642-16438-5_9}
Blomqvist, E., Presutti, V., Daga, E., Gangemi, A.: Experimenting with extreme design. In: Cimiano, P., Pinto, H.S. (eds.) Knowledge Engineering and Management by the Masses. pp. 120--134. Springer Berlin Heidelberg, Berlin, Heidelberg (2010)

\bibitem{blomqvist2005patterns}
Blomqvist, E., Sandkuhl, K.: Patterns in ontology engineering: Classification of ontology patterns. In: Proceedings of the Seventh International Conference on Enterprise Information Systems - Volume 3: ICEIS,. pp. 413--416. INSTICC, SciTePress (2005). \doi{10.5220/0002518804130416}

\bibitem{blomqvist2012ontology}
Blomqvist, E., Seil~Sepour, A., Presutti, V.: Ontology testing-methodology and tool. In: Knowledge Engineering and Knowledge Management: 18th International Conference, EKAW 2012, Galway City, Ireland, October 8-12, 2012. Proceedings 18. pp. 216--226. Springer (2012)

\bibitem{brown2020}
Brown, T., Mann, B., Ryder, N., Subbiah, M., Kaplan, J.D., Dhariwal, P., Neelakantan, A., Shyam, P., Sastry, G., Askell, A., Agarwal, S., Herbert-Voss, A., Krueger, G., Henighan, T., Child, R., Ramesh, A., Ziegler, D., Wu, J., Winter, C., Hesse, C., Chen, M., Sigler, E., Litwin, M., Gray, S., Chess, B., Clark, J., Berner, C., McCandlish, S., Radford, A., Sutskever, I., Amodei, D.: Language models are few-shot learners. In: Larochelle, H., Ranzato, M., Hadsell, R., Balcan, M., Lin, H. (eds.) Advances in Neural Information Processing Systems. vol.~33, pp. 1877--1901. Curran Associates, Inc. (2020), \url{https://proceedings.neurips.cc/paper_files/paper/2020/file/1457c0d6bfcb4967418bfb8ac142f64a-Paper.pdf}

\bibitem{chu2023survey}
Chu, Z., Chen, J., Chen, Q., Yu, W., He, T., Wang, H., Peng, W., Liu, M., Qin, B., Liu, T.: A survey of chain of thought reasoning: Advances, frontiers and future. arXiv preprint arXiv:2309.15402  (2023)

\bibitem{neongpt2024}
Fathallah, N., Das, A., De~Giorgis, S., Poltronieri, A., Haase, P., Kovriguina, L.: Neon-gpt: A large language model-powered pipeline for ontology learning. In: Proceedings of the ESWC2024 Special Track: Large Language Models for Knowledge Engineering (to appear) (2024)

\bibitem{fathallah2024llms4life}
Fathallah, N., Staab, S., Algergawy, A.: Llms4life: Large language models for ontology learning in life sciences. (2024)

\bibitem{Fernandez1997}
Fern\'{a}ndez, M., G\'{o}mez-P\'{e}rez, A., Juristo, N.: Methontology: from ontological art towards ontological engineering. In: Proceedings of the AAAI97 Spring Symposium Series on Ontological Engineering (1997)

\bibitem{LLM4KGs2024}
Frey, J., Meyer, L.P., Brei, F., Grunder-Fahrer, S., Martin, M.: Assessing the evolution of llm capabilities for knowledge graph engineering in 2023. In: Proceedings of the ESWC2024 Special Track: Large Language Models for Knowledge Engineering (to appear) (2024)

\bibitem{gangemi2005ontology}
Gangemi, A.: Ontology design patterns for semantic web content. In: The Semantic Web--ISWC 2005: 4th International Semantic Web Conference, ISWC 2005, Galway, Ireland, November 6-10, 2005. Proceedings 4. pp. 262--276. Springer (2005)

\bibitem{garijo2024llms}
Garijo, D., Poveda-Villalón, M., Amador-Dom{\'\i}nguez, E., Wang, Z., Gar{\'\i}a-Castro, R., Corcho, O.: Llms for ontology engineering: A landscape of tasks and benchmarking challenges. In: The Semantic Web – ISWC 2024: 23rd International Semantic Web Conference. Baltimore, MD, USA (November 11--15 2024), proceedings of the 23rd International Semantic Web Conference (ISWC 2024)

\bibitem{hanna2023comparative}
Hanna, E., Levic, A.: {Comparative Analysis of Language Models: Hallucinations in ChatGPT: Prompt Study}. Master's thesis, Linnaeus University (2023)

\bibitem{he24deeponto}
He, Y., Chen, J., Dong, H., Horrocks, I., Allocca, C., Kim, T., Sapkota, B.: Deeponto: A python package for ontology engineering with deep learning. (To appear in the Semantic Web Journal)  (2024)

\bibitem{keet2024roles}
Keet, C.M., Khan, Z.C.: On the roles of competency questions in ontology engineering. In: International Conference on Knowledge Engineering and Knowledge Management. pp. 123--132. Springer (2024)

\bibitem{khot2022decomposed}
Khot, T., Trivedi, H., Finlayson, M., Fu, Y., Richardson, K., Clark, P., Sabharwal, A.: Decomposed prompting: {A} modular approach for solving complex tasks. In: The Eleventh International Conference on Learning Representations, {ICLR} 2023, Kigali, Rwanda, May 1-5, 2023 (2023)

\bibitem{lippolisontogenia}
Lippolis, A.S., Ceriani, M., Zuppiroli, S., Nuzzolese, A.G.: {Ontogenia: Ontology Generation with Metacognitive Prompting in Large Language Models}. In: Poster and demos track, Satellite proceedings of ESWC2024 (to appear) (2024)

\bibitem{mateiu2023ontology}
Mateiu, P., Groza, A.: Ontology engineering with large language models. In: 2023 25th International Symposium on Symbolic and Numeric Algorithms for Scientific Computing (SYNASC). pp. 226--229. IEEE (2023)

\bibitem{peroni2016simplified}
Peroni, S.: A simplified agile methodology for ontology development. In: OWL: Experiences and Directions--Reasoner Evaluation, pp. 55--69. Springer (2016)

\bibitem{PluEscobarTrouillez2024}
Plu, J., Escobar, O.M., Trouillez, E., Gapin, A., Troncy, R.: A comprehensive benchmark for evaluating llm-generated ontologies. In: The Semantic Web – ISWC 2024: 23rd International Semantic Web Conference, Baltimore, MD, USA, November 11–15, 2024, Proceedings (2024)

\bibitem{poveda2014oops}
Poveda-Villal{\'o}n, M., G{\'o}mez-P{\'e}rez, A., Su{\'a}rez-Figueroa, M.C.: {OOPS! (OntOlogy Pitfall Scanner!): An On-line Tool for Ontology Evaluation}. International Journal on Semantic Web and Information Systems (IJSWIS)  \textbf{10}(2),  7--34 (2014)

\bibitem{POVEDAVILLALON2022104755}
Poveda-Villalón, M., Fernández-Izquierdo, A., Fernández-López, M., García-Castro, R.: Lot: An industrial oriented ontology engineering framework. Engineering Applications of Artificial Intelligence  \textbf{111},  104755 (2022). \doi{https://doi.org/10.1016/j.engappai.2022.104755}

\bibitem{presutti2009extreme}
Presutti, V., Daga, E., Gangemi, A., Blomqvist, E.: extreme design with content ontology design patterns. In: Proc. Workshop on Ontology Patterns. pp. 83--97. CEUR-WS (2009)

\bibitem{RebboudLisenaTroncy2024}
Rebboud, Y., Lisena, P., Tailhardat, L., Troncy, R.: Benchmarking llm-based ontology conceptualization: A proposal. In: The Semantic Web – ISWC 2024: 23rd International Semantic Web Conference, Baltimore, MD, USA, November 11–15, 2024, Proceedings (2024)

\bibitem{saeedizade2024navigating}
Saeedizade, M.J., Blomqvist, E.: Navigating ontology development with large language models. In: European Semantic Web Conference. pp. 143--161. Springer (2024)

\bibitem{shimizu_modular_2022}
Shimizu, C., Hammar, K., Hitzler, P.: Modular ontology modeling. Semantic Web  \textbf{14}(3),  459--489 (2023). \doi{10.3233/SW-222886}

\bibitem{Suárez-Figueroa2012}
Su{\'a}rez-Figueroa, M.C., G{\'o}mez-P{\'e}rez, A., Fern{\'a}ndez-L{\'o}pez, M.: The neon methodology for ontology engineering. In: Su{\'a}rez-Figueroa, M.C., G{\'o}mez-P{\'e}rez, A., Motta, E., Gangemi, A. (eds.) Ontology Engineering in a Networked World, pp. 9--34. Springer Berlin Heidelberg, Berlin, Heidelberg (2012). \doi{10.1007/978-3-642-24794-1_2}

\bibitem{NeOnBook}
Su{\'a}rez-Figueroa, M., G{\'o}mez-P{\'e}rez, A., Motta, E., Gangemi, A. (eds.): Ontology Engineering in a Networked World. Springer (2012)

\bibitem{TsanevaHerwantoSabou2024}
Tsaneva, S., Herwanto, G.B., Sabou, M.: Benchmarking ontology validation capabilities of llms. In: The Semantic Web – ISWC 2024: 23rd International Semantic Web Conference, Baltimore, MD, USA, November 11–15, 2024, Proceedings (2024)

\bibitem{wang2022}
Wang, B., Min, S., Deng, X., Shen, J., Wu, Y., Zettlemoyer, L., Sun, H.: Towards understanding chain-of-thought prompting: An empirical study of what matters. In: The 61st Annual Meeting Of The Association For Computational Linguistics (2023)

\bibitem{wang2023}
Wang, Y., Zhao, Y.: Metacognitive prompting improves understanding in large language models. In: Duh, K., Gomez, H., Bethard, S. (eds.) Proceedings of the 2024 Conference of the North American Chapter of the Association for Computational Linguistics: Human Language Technologies (Volume 1: Long Papers). pp. 1914--1926. Association for Computational Linguistics, Mexico City, Mexico (Jun 2024). \doi{10.18653/v1/2024.naacl-long.106}, \url{https://aclanthology.org/2024.naacl-long.106}

\bibitem{wei2022}
Wei, J., Wang, X., Schuurmans, D., Bosma, M., Xia, F., Chi, E., Le, Q.V., Zhou, D., et~al.: Chain-of-thought prompting elicits reasoning in large language models. In: Advances in Neural Information Processing Systems. pp. 24824--24837 (2022)

\end{thebibliography}
%






 \pagebreak
\section{Appendix}
\label{sec:appendix}

\subsection{Examples of Evaluation Metrics}
\label{sec:Metrics}

\begin{figure}
    \centering
    \includegraphics[width=0.7\linewidth]{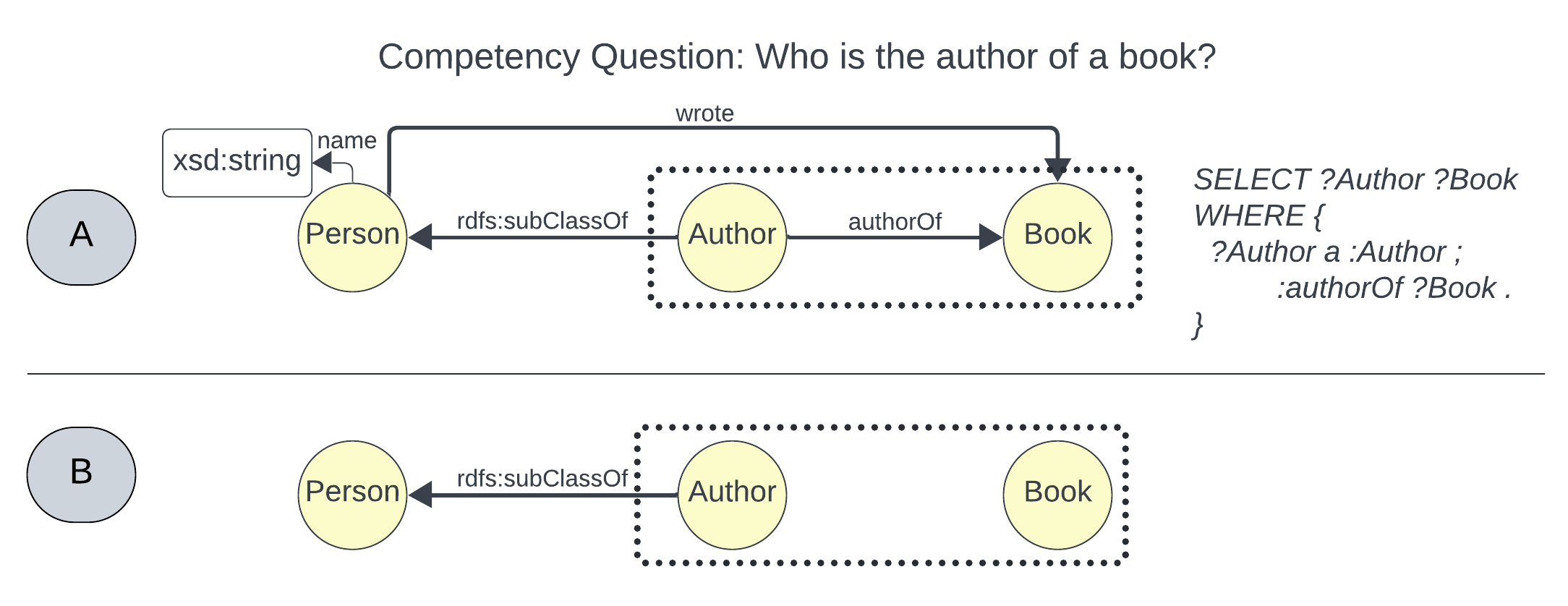}
\caption{Analysis of an ontology for a CQ. Part A shows a correctly modelled CQ, ensuring all necessary elements for a SPARQL query are present, but contain superfluous elements. Part B shows a minor issue where a data property is missing.}

    \label{fig:CQs}
\end{figure}

\textbf{Modelled CQ:} In Figure \ref{fig:CQs}, part A (top), a CQ such as ``Who is the author of a book?'' is shown to be correctly modelled, as all the elements required to write a SPARQL query are present in the model (dotted box).

\textbf{Minor Issue:} In part B, for the same CQ, an object property to connect Book and Author (or either connecting it to Person or a new data property) is missing. However, adding ``authorOf'' would resolve the issue. This is considered a minor error, as the ontology modeller can easily add this element.

\textbf{Superfluous Elements:} In Figure \ref{fig:CQs}, part A, for the given CQ, there is a class ``Person'' and properties such as ``name'' or ``wrote'' that are not used in the SPARQL query (top-right corner). Since they (outside of the dotted box) do not appear in the SPARQL, they are referred to as superfluous elements.

Each metric evaluates a specific aspect of the LLM-generated ontology, as no single metric is sufficient to provide a comprehensive assessment. For instance, Llama produces many superfluous elements, hence it also has a reasonable likelihood of randomly modelling certain simple CQs correctly. By penalizing superfluous elements, we prevent LLMs from generating too many unnecessary classes and properties; however, this approach may also discourage the generation of taxonomies or more generalized modelling, which is a disadvantage. 
By combining all metrics, we can achieve an holistic evaluation of ontology generation.

\subsection{Discussion on LLMs}
\textbf{Comparison of LLMs.}
By comparing the results of the LLMs in Section \ref{sec:results}, we can observe that \texttt{o1-preview} is the most promising LLM for ontology generation and GPT-4 may serve as its replacement. However, while Llama exhibits similar performance to GPT-4 in correctly modelling CQs, it produces a considerable number of superfluous elements and critical OOPS! pitfalls and performs the worst according to the expert evaluation. Thus, it may still not be possible to create an ontology engineering co-pilot using open models, and hence the issue of cost and the risks related to the black-box nature of proprietary models, remain. 

\textbf{Context size.}
Comparing the scores of Memoryless CQbyCQ with other techniques shows that reducing the input context size of \textit{GPT-4} significantly enhances the output score of this model despite producing some superfluous elements. These results highlight a trade-off between computational efficiency, cost, and output quality. If quality is a priority but resources are limited, Memoryless CQbyCQ balances efficiency and speed, though it may require additional post-processing to refine the output.
\subsection{Limitations }

\textbf{Data leakage assessment} A critical consideration in our ontology selection criteria and dataset publication is the potential for data leakage \cite{balloccu2024leakcheatrepeatdata}, due to the online accessibility of some of this material. Such exposure introduces risks, namely the likelihood of evaluating LLMs on the same datasets they were trained on. Also, the dataset published in this paper as a gold standard might be used in the training data of future LLMs. Our selected ontologies are published with their CQs and user stories separately from their ontology models, which can mitigate these risks. The ontologies from the courses (Music, Hospital and Theatre) date back to 2008-2009, but only the stories and CQs have been published online, not the solutions in OWL. Moreover, to further protect our dataset being used in LLMs training data, we provide it in a zipped folder that is secured with a password, ensuring controlled access and reducing the risk of data leakage for future research in this field.

\textbf{Additional Limitations }
This study's scope is to see to what extent LLMs can serve in generating ontologies. There is a need for more evaluation criteria to assess the utility and generalizability of generated ontologies. Our ontologies, drawn from different domains, are primarily single-module, limiting their reusability across different systems. Moreover, our methods do not adequately prevent or eliminate superfluous elements, as shown by model output repetitions. Our Memoryless CQbyCQ approach, which reduces context size to enhance performance and reduce cost, is not suitable for history-dependent modelling tasks that heavily depend on the communication history with LLMs. This technique is more applicable to methods where each part of the ontology development can be performed independently in small parts and where there is an effective strategy for merging these partial solutions by human-in-the-loop integration. Additionally, further tests are needed to mitigate the potential leakage effect and bias in LLMs. One way to achieve this is by applying the method to new use cases in different domains. Future work will involve mitigating the limitations, for example, through manual re-engineering of the generated ontology draft. Potentially the latter can be supported by user interfaces, e.g., in the form of a plugin.

\end{document}